\documentclass{article}
\usepackage{ijcai26}

\usepackage{times}
\usepackage{soul}
\usepackage{url}
\usepackage[utf8]{inputenc}
\usepackage[small]{caption}
\usepackage{graphicx}
\usepackage{amsmath}
\usepackage{amsthm}
\usepackage{booktabs}
\usepackage{algorithm}
\usepackage{algorithmic}
\usepackage[switch]{lineno}
\usepackage{mathtools}

\usepackage{subcaption}
\usepackage{xcolor}
\usepackage{array}
\usepackage{multirow}
\usepackage[inline]{enumitem}
\usepackage{rotating}
\usepackage{tipa}
\usepackage[nohyperlinks, nolist]{acronym} 

\usepackage{siunitx}
\usepackage{amssymb}
\usepackage{tabularx}
\usepackage[hidelinks]{hyperref}
\usepackage{cleveref}

\usepackage{tikz}
\usepackage{pgfplots}
\pgfplotsset{compat=1.18}

\newcommand{\citet}[2]{#1~\cite{#2}}

\usetikzlibrary{shapes.geometric}

\title{Faithfulness Without Anchors: A Position Paper on the Structural Misalignments of Explainable AI Faithfulness Metrics for Feature Importance\thanks{Presented at the IJCAI-ECAI 2026
Workshop on Explainable
Artificial Intelligence (XAI)}}

\author{
   Marco Zullich$^1$\and
    Emily Schiller$^{2,3}$\and
   Ivan Gentile$^4$\and
   David Dembinsky$^{5,6}$ \\
   Adriano Lucieri$^{6}$\and
   Kilian Göller$^{7}$\and
   Steffen Seitz$^{7,8}$
  \affiliations
   $^1$TU Delft, %Faculty of Technology, Policy and Management, Delft University of Technology, Delft, the Netherlands\\
    $^2$XITASO GmbH, %IT \& Software Solutions, Augsburg, Germany\\
    $^3$University College Cork, %The Artificial Intelligence and Cognitive Load Research Lab, University College Cork, Cork, Ireland\\
   $^4$IFAB Foundation,
   $^5$RPTU University Kaiserslautern-Landau,
    $^6$DFKI, 
   $^7$TU Dresden,
   $^8$SECAI
   \emails
   All authors in no particular order. \\Correspondence: m.zullich@tudelft.nl
}

\begin{document}

\maketitle

\begin{abstract}
    Explainable Artificial Intelligence (XAI) tools have often been presented as methods to increase trust in machine learning models by making aspects of their algorithms more human-understandable.
    However, evidence accumulated over the last decade points to an overall failure of this objective, instead promoting overreliance on incorrect predictions, which can subsequently lead to algorithmic aversion.
    Rather, XAI appears to function primarily as a debugging tool, useful for obtaining anecdotal insights into, for example, features learned by the model.
    One of the main issues underlying this limitation is that XAI methods often oversimplify the inner workings of the model, resulting in misalignment between the explanation and the model itself—a behavior referred to as \emph{unfaithfulness} or \emph{infidelity}.
    \emph{Faithfulness} can be interpreted as the degree of approximation between an explanation and the model.
    Computing faithfulness in a rigorous manner would thus make it possible to quantify one of the various elements of trustworthiness of an explanation.
    However, existing metrics present significant limitations, making this quantification either infeasible or computationally expensive.
    In this position paper, we outline five major shortcomings of current faithfulness metrics and propose four possible avenues for research to bolster their reliability.
    \end{abstract}

\begin{acronym}
    \acro{AI}[AI]{artificial intelligence}
    \acro{ANN}{artificial neural network}
    \acro{AOPC}{area over the perturbation curve}
    \acro{AUPC}{area under the perturbation curve}
    \acro{ECE}{expected calibration error}
    \acro{FI}[FI]{feature importance}
    \acro{LERF}[LeRF]{least relevant first}
    \acro{MORF}[MoRF]{most relevant first}
    \acro{OOD}[OoD]{out-of-distribution}
    \acro{ROAR}[ROAR]{remove-and-retrain}
    \acro{UE}[UE]{uncertainty estimation}
    \acro{XAI}[XAI]{explainable AI}
\end{acronym}

\section{Introduction}\label{sec:intro}

With the increasing power of \ac{AI} models, their internal algorithmic processes have become more difficult to interpret, as exemplified in the notorious interpretability-accuracy trade-off \cite{james2013introduction}.
This opacity conflicts with transparency requirements such as the GDPR ``Right to Explanation'' \cite{goodman2017european}.
\Ac{XAI} has emerged as a field to enable transparency of \emph{black-box} models by providing \emph{explanations} that simplify \emph{black-box} models for human understanding.
Haufe \textit{et al.}~\shortcite{haufe2026explainable} indicate ideal goals of \ac{XAI}, including the recognition of model and data diagnostics (i.e., discovering incorrect model behaviors, such as ``Clever Hans effects'' \cite{lapuschkin2019unmasking}, or bias in datasets), the discovery of unknown relationships between variables, and the identification of intervention targets for modifying said relationships.
However, evidence from roughly ten years of \ac{XAI} research shows these tools often fall short on these promises, as recently discussed by Afroogh \textit{et al.}~\shortcite{afroogh2026beyond}.
Beyond highlighting misalignment in explanation recipients, they report that \ac{XAI} can foster user overreliance on incorrect \ac{AI} predictions that would otherwise be questioned. In safety-critical applications, this may cause serious consequences and potential algorithmic aversion.
Both Haufe \textit{et al.}~\shortcite{haufe2026explainable} and Afroogh \textit{et al.}~\shortcite{afroogh2026beyond} indicate \emph{infidelity} (or \emph{unfaithfulness}) of \ac{XAI} outputs to the original model as a source of said downfalls.

However, these works do not examine the technical issues of faithfulness metrics, nor explore faithfulness beyond \ac{FI}.
In their survey on the evaluation of \ac{XAI}, Dembinsky \textit{et al.}~\shortcite{dembinsky2026unifying} cite 19 faithfulness metrics, applicable to different types of \ac{XAI}: \ac{FI}, concepts, example explanations, white-box surrogates, and natural language explanations. Faithfulness for \ac{FI} is by far the type with the most approaches, driven also by the fact that \ac{FI} is currently the most common \ac{XAI} task \cite{saarela2024recent}.
Nevertheless, techniques for the computation of faithfulness, which mostly rely on (iterative) feature removal, are problematic at best:
they are unreliable---e.g., due to issues with \ac{OOD} data, require arbitrary baselines, or require re-training phases which alter the nature of the original model.
Some other fidelity metrics do not even measure proper faithfulness, or conflate it with other \ac{XAI} properties.
This unreliability raises the question whether \acp{FI} can be truly considered \emph{unfaithful}, or whether faithfulness metrics themselves are dubious.

In this position paper, we analyze recent studies on faithfulness metric unreliability, arguing that modern \ac{XAI} should prioritize more trustworthy approaches. The field should move beyond established but unreliable metrics like pixel flipping \cite{bach2015pixel} and \ac{ROAR} \cite{hooker2019benchmark} toward techniques considering
\begin{enumerate*}[label=(\roman*)]
    \item what is the real goal behind faithfulness,
    \item the model's \ac{UE} capabilities, and
    \item expected properties of \ac{FI} coefficients with respect to model predictions.
\end{enumerate*}
We conclude with research avenues for advancing faithfulness evaluation in \ac{FI}, restricting our claims to this domain given the substantial available research.

%This paper is organized as follows: \Cref{sec:background} introduces \ac{FI} and its evaluation approaches. \Cref{sec:faith-issues} outlines five major shortcomings of current faithfulness computation. \Cref{sec:proposals} presents four recommendations for more trustworthy faithfulness research. \Cref{sec:related} reviews related criticisms of current metrics, followed by a critical discussion in \Cref{sec:discussion}.

\section{Background}\label{sec:background}

% This section provides the necessary background, introducing \ac{FI}, then faithfulness, and finally criticisms of current computation approaches.

\subsection{Feature Importance}\label{sec:background-fi}

\ac{FI} is one of the foundational types of \ac{XAI} \cite{dembinsky2026unifying}.
Let us consider a model $f:\mathbb{R}^d\rightarrow\mathcal{Y}$, where $\mathcal{Y}$ is the output space.
We can define a \ac{FI} technique as a mapping $e:\mathbb{R}^d\rightarrow\mathbb{R}^d$ whose output we term \emph{explanation}.
The \ac{FI} explanation provides a value (which we can term importance \emph{coefficient} or \emph{score}) for each feature\footnote{
%Usually, \ac{FI} assigns one value per feature; however, there exist approaches that consider arbitrary orders of feature interactions: their output space is $\mathbb{R}^{d^k}$, $k$ being the highest order of feature interaction contemplated by the model \cite{fumagalli2023shap}.
%Additionally, other \ac{FI} approaches consider grouping of raw features, as in the case of LIME's original approach for images \cite{ribeiro2016should}, or generates the explanations in a different output space, as in the case of SoundLIME \cite{mishra2020reliable}, where the \ac{FI} is produced in three different domains.
For simplicity, we consider only \ac{FI}s that generate one importance value per raw feature.
This is not necessarily true, e.g., when considering feature interaction \cite{fumagalli2023shap}, supersets of features \cite{ribeiro2016should}, or explanations in different domains than the input's \cite{mishra2020reliable}.
}, each value indicating the \emph{importance} of the feature in the model's predictive dynamics.
\ac{FI} can be \emph{global} (thus, the importance coefficients are valid for the model as a whole) or \emph{local} (i.e., the importance scores are valid only for one specific point of the input space).
Importance scores with a high magnitude indicate that the model relies greatly on the corresponding features for outputting the prediction; the sign instead indicates whether the feature contributes in increasing the model prediction (if positive) or in decreasing it (if negative).
Some techniques might only present positive coefficients: for instance Permutation Feature Importance \cite{fisher2019all} only outputs importance in term of magnitude, while other techniques, such as Grad-CAM \cite{selvaraju2020grad} suppresses negative importance scores for simplifying the interpretation of the explanation.
A faithful \ac{FI} should correctly indicate how important individual features are in the model's predictive dynamics.
In the case of some \ac{FI} techniques, such as SHAP \cite{lundberg2017unified}, this is often defined in terms of \emph{feature removal}:
%The desiderata for the importance scores is that 
the deletion of an important feature (leaving the others intact) should cause a change in the model prediction, and that this change should be proportional to the importance coefficient.
However, other tools, such as gradient-based \ac{FI} for neural networks just report (aggregations of) gradients as scores, without referring to feature removal.
\ac{FI} can also be called \emph{feature attribution}; when working with image data, these explanations can also be termed \emph{(importance) heatmaps} or \emph{saliency maps}.

\subsection{Evaluation of Feature Importance}

According to Nauta \textit{et al.}~\shortcite{nauta2023anecdotal}, explanations can be evaluated from the perspectives of their content, their presentations, or their users.
The assessment of the content can be further grouped in different categories---alignment between model and explanation (which the authors call \emph{correctness} and \emph{completeness} and that overall makes up the faithfulness \cite{dembinsky2026unifying}), continuity (robustness) of the explanation to perturbations (to input, output, or model), and compactness, among others. 
The presentation point of view assesses the mean through which the explanation is presented, regardless of its content.
The user viewpoint describes instead the conditions (context, previous knowledge, etc.) in which the user interacts with the explanation. A subgroup of the user category worth mentioning is \emph{coherence}, which other works also call \emph{plausibility}: the alignment between an explanation and human prior knowledge.

\subsubsection{Faithfulness in Feature Importance}\label{sec:faithfulness}

\begin{figure}
    \centering
    \scalebox{0.75}{
        \begin{tikzpicture}
\begin{axis}[
    % Axis style — y-axis aligned at x=0
    axis lines = left,
    axis line style = {-stealth, thick},
    % Labels
    xlabel = {Proportion of features removed},
    ylabel = {Model confidence},
    xlabel style = {font=\large, at={(axis description cs:0.15,-0.1)}, anchor=west},
    ylabel style = {font=\large, at={(axis description cs:-0.05,0.475)}, anchor=south},
    % Range: xmin=0 so y-axis sits exactly at x=0
    xmin = 0, xmax = 2.1,
    ymin = -0.05, ymax = 1.2,
    % No auto ticks
    xtick = \empty,
    ytick = \empty,
    % Size
    width  = 9cm,
    height = 7cm,
    clip = false,
]

% BLUE: steep early, asymptotic at end
\addplot[
    domain = 0:2,
    samples = 400,
    color = blue!70!black,
    line width = 1.8pt,
] {0.1 + 0.9 * exp(-3.5 * x)};

% RED: slow start, fast decay at end (time-reversed blue)
\addplot[
    domain = 0:2,
    samples = 400,
    color = red!70!black,
    line width = 1.8pt,
] {1 - 0.9 * (exp(-3.5*(2-x)) - exp(-7)) / (1 - exp(-7))};

% STRAIGHT LINE from (0,1) to (2, 0.1)
\addplot[
    domain = 0:2,
    samples = 2,
    color = black,
    line width = 1.2pt,
    dashed,
] {1 + (0.1 - 1)/2 * x};

% Dashed horizontal asymptote at y = 0.1
% \draw[dashed, gray!50, thin]
%    (axis cs:0, 0.1) -- (axis cs:2.5, 0.1);

% Y-axis: tick + label C(0) at y=1
\draw[thick] (axis cs:0, 1) -- (axis cs:-0.06, 1);
\node[anchor=east, font=\normalsize] at (axis cs:-0.05, 1) {$\hat{y}_0$};

% Y-axis: tick + label C(1) at y=0.1
\draw[thick] (axis cs:0, 0.1) -- (axis cs:-0.06, 0.1);
\node[anchor=east, font=\normalsize] at (axis cs:-0.05, 0.1) {$\hat{y}_1$};

% X-axis: tick + label at x=2
\draw[thick] (axis cs:2, 0) -- (axis cs:2, -0.04);
\node[anchor=north, font=\normalsize] at (axis cs:2, -0.05) {$1$};

\node[anchor=north, font=\normalsize] at (axis cs:0, -0.05) {$0$};

\node[] at (0.25, 0.25) {\ac{MORF}};
\node[] at (1.50, 0.95) {\ac{LERF}};

\end{axis}
\end{tikzpicture}
    }
    \caption{Example of possible perturbation curves obtained in a classification setting for an incremental deletion faithfulness evaluation.
    The model starts by predicting a confidence value $\hat{y}_0$ for the predicted class on the unperturbed data.
    As more features are removed, the model confidence drops until reaching a minimum $\hat{y}_1$ when the data point is completely perturbed.
    The curve depicted in blue shows an optimal \ac{MORF} confidence trend, while the red curve is expected to be attained for a \ac{LERF} perturbation. 
    }
    \label{fig:perturbation-curve}
\end{figure}
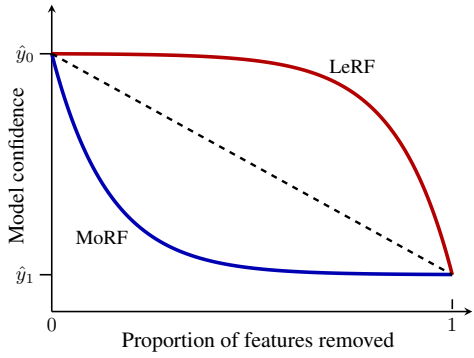

Faithfulness (also called \emph{fidelity} or \emph{correctness} in the literature) describes how truthful the explanation is to the original model $f$ \cite{nauta2023anecdotal}.
For \ac{FI}, this requires alignment between features the model truly considers important and those the explanation identifies as important.
In their survey, Dembinsky \textit{et al.}~\shortcite{dembinsky2026unifying} identify input perturbation as the primary methodology for evaluating \ac{FI} faithfulness. 
The rationale follows from the feature removal considerations presented in \Cref{sec:background-fi}: perturbing (i.e., \emph{removing}) important features should substantially change model predictions, with magnitude proportional to feature importance. The sign should be the opposite of the sign of the \ac{FI} output.
The authors distinguish between \emph{unguided} and \emph{guided} perturbation approaches.
Unguided approaches operate a perturbation without considering the hierarchy dictated by the \ac{FI} output.
Guided approaches, conversely, perturb a data point sequentially targeting features by importance, either most-to-least in \ac{MORF} or least-to-most in \ac{LERF}.
Nauta \textit{et al.}~\shortcite{nauta2023anecdotal} additionally distinguish between \emph{incremental deletion} or \emph{incremental addition} approaches: in the former, the features are sequentially perturbed starting from an unperturbed data point; in the latter the initial data point is completely perturbed, and the original features are \emph{revealed} in \ac{MORF} or \ac{LERF} order.
Plotting model output versus perturbation iteration yields a \emph{perturbation curve}.
In the case of classification, the model \emph{confidence} (probability value associated to the predicted class) is often used as output in the plot.
We showcase a possible example for classification in \Cref{fig:perturbation-curve}---\ac{MORF} removal causes a sharp decrease in confidence at first, while removing less important features has a small effect on the output; with \ac{LERF} removal the confidence drops at an increasing rate, proportional to the importance of features removed.
We point out that the behavior presented in \Cref{fig:perturbation-curve} is ideal---often, for reasons explored in the next sections, the situation can be very noisy.
Other factors, such as \emph{gradient starvation} \cite{pezeshki2021gradient}, can also modify the behavior of the curve:
if multiple features are associated with a given prediction, models may concentrate originally only on a small subset of these.
After perturbation of important features, the models may shift the focus on the other features, while still outputting the same prediction, modifying the \emph{ideal} scenario in \Cref{fig:perturbation-curve}.
To quantify faithfulness for an $(f, e)$ pair, the technique is applied over $n$ data points and aggregated via:
\begin{enumerate}[label=\roman*)]
    \item Correlation-based metrics: compute the Pearson's or Spearman's correlation, or Kendall's $\tau$, between the importance of the perturbed features and the opposite of the change in model prediction.
    A positive correlation should be observed.
    \item Area-based metrics: plot the model prediction over the importance of perturbed features and measure faithfulness as the \ac{AUPC} or \ac{AOPC} (or a combination of both \cite{knoll2025xaiev}). The metric can be reported either absolute or normalized. 
    Depending on the approach adopted---incremental deletion versus addition, \ac{MORF} versus \ac{LERF}---a highly faithful explanation should lead to a normalized area value close to 1 or 0 (if the metric is normalized).
\end{enumerate}

\section{Issues with existing faithfulness metrics}\label{sec:faith-issues}

%In this section, we outline major issues with existing approaches to computing faithfulness for \ac{FI}.

\subsection{Faithfulness computation is almost exclusively operated on classification}\label{sec:criticism-regression}

Most of the literature on faithfulness computation is limited to classification settings (e.g., \cite{bhatt_evaluatingandaggregating,zheng2024ffidelity}), while regression tasks remain insufficiently addressed. This limitation is not merely a gap in coverage but reflects incompatibilities between metric design assumptions and regression. Most metrics assume that removing (or inserting) important features should induce monotonic degradation (or improvement) in model confidence, specifically measured through classification scores~\cite{samek2016evaluating} such as softmax probabilities or logits. This assumption is well-grounded in classification, where the relevant quantity, class probability, is naturally bounded between 0 and 1, and where degradation has a clear interpretation of decreased confidence. In regression, the target quantity itself is ambiguous: should faithfulness measure changes in the predicted value or in deviation from the ground truth~\cite{demetriou2025study}? These probe fundamentally different qualities. Because both quantities are typically unbounded, normalization becomes necessary but problematic: faithfulness scores become incomparable across datasets and even across instances within the same dataset, undermining their utility as evaluation metrics.  Additionally, perturbations may cause arbitrary shifts in predictions without clear monotonic relationships, making the ``degradation'' criterion ill-defined. Focusing on prediction error conflates model accuracy with explanation quality, measuring whether the model fails rather than whether the explanation faithfully represents the model's reasoning. Altogether, the literature reflects a deep bias of faithfulness metrics for \ac{FI} toward classification tasks that cannot be trivially adapted to regression contexts.

\subsection{Some faithfulness metrics conflate fidelity with other properties of explanations}\label{sec:criticism-conflation}

The very definition of faithfulness which we presented in \Cref{sec:faithfulness} is often misrepresented in some mainstream approaches for faithfulness computation presented in the literature.
We first note that sometimes the term \emph{correctness} is confused for \emph{plausibility}, as in the cases of Rony and Hassan \shortcite{rony2025between}, and Maathuis \textit{et al.}~\shortcite{maathuis2026designing}.
%Other papers, such as in the case of Lago \textit{et al.}~\shortcite{lago2026evaluating} conflate fidelity with the notions of robustness.

On a more foundational level, we notice that, additionally, some metrics for faithfulness conflate faithfulness itself with the notion of plausibility.
Since faithfulness is aimed at measuring the alignment between the explanation and the model, its evaluation should not depend on external notions such as human-generated labels or ground-truth annotations.
In line with the observations on fidelity computation from Jacovi and Goldberg \shortcite{jacovi2020towards}, we posit that faithfulness should be assessed only on the basis of the model internals.
Similarly, Dembinsky et al.~\shortcite{dembinsky2026unifying} argue that ground-truth dataset evaluation assesses faithfulness only when a uniquely defined rationale can be assumed beyond reasonable doubt; otherwise, it merely evaluates plausibility.
However, some highly-adopted approaches for faithfulness computation target changes in performance metrics of the model instead of prediction values.
This is the case, e.g., for ROAR \cite{hooker2019benchmark} and ROAD \cite{rong2022consistent}.
ROAR and ROAD measure faithfulness in terms of performance difference after perturbation (and re-training for the former).
Measuring faithfulness, e.g., in terms of accuracy drops is probing the alignment between explanation and ground truth, not between explanation and model.

% maybe new subsection?
Additionally, since ROAR requires a re-training of the model on perturbed input, we posit that this metric is not even assessing the quality of the explanations according to the original model, but according to a new model, which can be completely distinct in its internals from the original---an issue which Dembinsky \textit{et al.}~\shortcite{dembinsky2026unifying} call \emph{different contextuality}.
For these reasons, we will avoid including ROAR in future discussions, even though it is often cited as being a state-of-the-art faithfulness metric \cite{song2023pitfalls,klein2024navigating}.

\subsection{Perturbation as a stand-in for feature removal requires arbitrary baselines}\label{sec:criticism-baseline}

As stated in \Cref{sec:faithfulness}, perturbations are involved in the faithfulness computation procedure as a stand-in for feature removal, which is often impossible to reproduce with most machine learning models unless a retraining procedure is operated.
The main problem with this methodology is that there does not exist a single best solution for reproducing feature removal---the baseline for this procedure is thus arbitrary and it is thus impossible to define what perturbation would encode this \emph{missingness of information}.
The problem is equivalent to the one faced by \ac{FI} tools that work by feature removal, the most important one being SHAP.
Strumfels \textit{et al.}~\shortcite{sturmfels2020visualizing} already ran an extensive study on the effect of the baseline selection (constant baseline, blurring, random uniform and gaussian) for image-based \ac{FI} generation, which led to the conclusion of there not being a ``best'' solution.
For what concerns faithfulness computation, different approaches are presented in the literature and largely depend on the data type:
in image data, the most common approach is \emph{fixed imputation}, where a patch of a uniform color (e.g., white, gray, black, or the dataset mean) is applied to the image to \emph{occlude} important features \cite{samek2016evaluating,sundararajan2017axiomatic}; other approaches use blurring or noising \cite{zheng2024ffidelity} or use generative models for image inpainting \cite{agarwal2020explaining}.
For text data, hard token deletion can be applied \cite{deyoung2020eraser}, or tokens can be replaced with specific mask tokens \cite{atanasova2020diagnostic}, fixed embeddings (e.g., zero-embedding \cite{li2016visualizing}) or with tokens generated by another language model \cite{kim2020interpretation,wu2021polyjuice}.
In tabular data, common approaches include fixed imputation with a constant or marginal baseline or noising.

\subsection{Many perturbation-based approaches cause \ac{OOD} issues with models}\label{sec:criticism-ood}

Besides the difficulty with identifying what information missingness means for a specific model or type of data, the search for \emph{good} perturbation baselines is also driven by another noticeable issue with \acp{ANN}: their overconfidence on \ac{OOD} data.
Indeed, these models have been largely known to output random, overconfident predictions when presented with data which are too \emph{distant} from their original training data distribution \cite{nguyen2015deep}.
All the perturbation strategies insofar presented are somewhat sensitive to the \ac{OOD} issue \cite{hase2021out}, and some \cite{rong2022consistent,zheng2024ffidelity} have specifically been developed to tackle this problem.
Another connected issue is that classification \acp{ANN}, unless explicitly trained on data with missing information, or correctly calibrated on the uncertainty profile \cite{guo2017calibration}, these models do not possess the ability to say ``\emph{I do not know}'' (expressed as a flat output distribution).
The way the \ac{OOD} issue reflects on the perturbation curve is the fact that the model response becomes erratic, with highly-perturbed data often having a higher confidence than the previous iterations.
We display a real-life example in \Cref{fig:catchall}.
However, the faithfulness metrics introduced in \Cref{sec:faithfulness} all work under the assumption that the perturbation curve is monotonically decreasing, which clearly conflicts from the ``clean'' examples from \Cref{fig:perturbation-curve}.

The main approaches for combating the \ac{OOD} issue in faithfulness are:
\begin{enumerate*}[label=(\roman*)]
    \item augmentative training \cite{hase2021out} (inlcuding the F-fidelity framework \cite{zheng2024ffidelity}), where the training data is augmented by reproducing the perturbation patterns used in the faithfulness assessment, and
    \item generative inpainting \cite{agarwal2020explaining}, where the features are removed using an inpainting model that should still produce feasible data.
\end{enumerate*}
However, these two approaches do not solve the problem:
augmentative training only works until a reasonable level of perturbation.
High levels of perturbation will likely cause a complete erasing of the original features, which, in the case of supervised learning, will potentially force the model to learn an incorrect relationship between data and response.
%Even in the case of F-fidelity, where the perturbation is not strong enough to completely erase information, the model learns to predict with high confidence, even with heavy perturbation.
This potentially causes confident predictions even with a large portion of important features removed.
Generative inpainting causes absence of information in the image, which a non-calibrated model does not know how to interpret---a behavior that leads to overconfident \ac{OOD} predictions, thus impairing an approach which in principle should be valid.
Notice also that these approaches are mainly aimed at image data, and may not be easily translated to other data modalities.
Recent advances in \ac{XAI}-aware model calibration \cite{sridhar2026mcal,cai2026rethinking} seem to go in the right direction: address the model calibration, without proposing other workarounds for \ac{OOD} detection.

% \subsection{Both correlation-- and area--based metrics are actually misaligned from measuring faithfulness}
\subsection{The Formal Divergence of Correlation and Area-Based Metrics from Explanatory Faithfulness} \label{sec:criticism-formal-divergence}

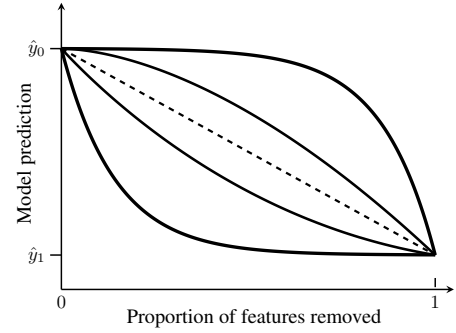
\begin{figure}[t]
    \centering
    \scalebox{0.7}{
        \begin{tikzpicture}
\begin{axis}[
    % Axis style — y-axis aligned at x=0
    axis lines = left,
    axis line style = {-stealth, thick},
    % Labels
    xlabel = {Proportion of features removed},
    ylabel = {Model prediction},
    xlabel style = {font=\large, at={(axis description cs:0.15,-0.1)}, anchor=west},
    ylabel style = {font=\large, at={(axis description cs:-0.05,0.475)}, anchor=south},
    % Range: xmin=0 so y-axis sits exactly at x=0
    xmin = 0, xmax = 2.1,
    ymin = -0.05, ymax = 1.2,
    % No auto ticks
    xtick = \empty,
    ytick = \empty,
    % Size
    width  = 9cm,
    height = 7cm,
    clip = false,
]

% BLUE: steep early, asymptotic at end
\addplot[
    domain = 0:2,
    samples = 400,
    %color = blue!70!black,
    line width = 1.8pt,
] {0.1 + 0.9 * exp(-3.5 * x)};

% RED: slow start, fast decay at end (time-reversed blue)
\addplot[
    domain = 0:2,
    samples = 400,
    %color = red!70!black,
    line width = 1.8pt,
] {1 - 0.9 * (exp(-3.5*(2-x)) - exp(-7)) / (1 - exp(-7))};

\addplot[domain=0:2, samples=400, color=black, line width=1.4pt]
    {1 - 0.9*x + 0.9*(x/2)^1.7};

\addplot[domain=0:2, samples=400, color=black, line width=1.4pt]
    {1 - 0.9*(x/2)^1.7};

% STRAIGHT LINE from (0,1) to (2, 0.1)
\addplot[
    domain = 0:2,
    samples = 2,
    color = black,
    line width = 1.2pt,
    dashed,
] {1 + (0.1 - 1)/2 * x};

% Dashed horizontal asymptote at y = 0.1
% \draw[dashed, gray!50, thin]
%    (axis cs:0, 0.1) -- (axis cs:2.5, 0.1);

% Y-axis: tick + label C(0) at y=1
\draw[thick] (axis cs:0, 1) -- (axis cs:-0.06, 1);
\node[anchor=east, font=\normalsize] at (axis cs:-0.05, 1) {$\hat{y}_0$};

% Y-axis: tick + label C(1) at y=0.1
\draw[thick] (axis cs:0, 0.1) -- (axis cs:-0.06, 0.1);
\node[anchor=east, font=\normalsize] at (axis cs:-0.05, 0.1) {$\hat{y}_1$};

% X-axis: tick + label at x=2
\draw[thick] (axis cs:2, 0) -- (axis cs:2, -0.04);
\node[anchor=north, font=\normalsize] at (axis cs:2, -0.05) {$1$};

\node[anchor=north, font=\normalsize] at (axis cs:0, -0.05) {$0$};

\end{axis}
\end{tikzpicture}
    }
    \caption{Example of perturbation curves whose Spearman's correlation coefficient is always -1. The dotted line represents a case in which Pearson's correlation coefficient is -1. In either case (\ac{MORF} or \ac{LERF}), this would lead to a perfect faithfulness, although these configurations clearly represent different cases.
    In the \ac{MORF} case, the lower curves would represent an ideal scenario, whereby the most important features have a higher impact on the model prediction; the dotted line and the two upper curves represent instead an unsound scenario for faithfulness, with highly-important feature having close to no effect on the prediciton, while features with a lower importance contribute more.
    This shows the fault in correlation coefficients as a faithfulness metric.}
    \label{fig:correlation}
\end{figure}

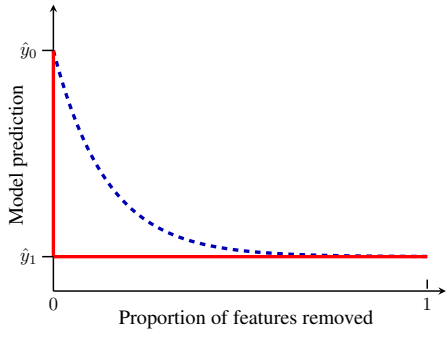
\begin{figure}[t]
    \centering
    \scalebox{0.7}{
        \begin{tikzpicture}
\begin{axis}[
    % Axis style — y-axis aligned at x=0
    axis lines = left,
    axis line style = {-stealth, thick},
    % Labels
    xlabel = {Proportion of features removed},
    ylabel = {Model prediction},
    xlabel style = {font=\large, at={(axis description cs:0.15,-0.1)}, anchor=west},
    ylabel style = {font=\large, at={(axis description cs:-0.05,0.475)}, anchor=south},
    % Range: xmin=0 so y-axis sits exactly at x=0
    xmin = 0, xmax = 2.1,
    ymin = -0.05, ymax = 1.2,
    % No auto ticks
    xtick = \empty,
    ytick = \empty,
    % Size
    width  = 9cm,
    height = 7cm,
    clip = false,
]

% BLUE: steep early, asymptotic at end
\addplot[
    domain = 0:2,
    samples = 400,
    color = blue!70!black,
    dashed,
    line width = 1.8pt,
] {0.1 + 0.9 * exp(-3.5 * x)};

% Solid red vertical line at x=0 from C(1) to C(0)
\draw[red, line width=1.8pt] (axis cs:0, 0.1) -- (axis cs:0, 1);

% Solid red horizontal line from (0, C(1)) to (2, C(1))
\draw[red, line width=1.8pt] (axis cs:0, 0.1) -- (axis cs:2, 0.1);

% Dashed horizontal asymptote at y = 0.1
% \draw[dashed, gray!50, thin]
%    (axis cs:0, 0.1) -- (axis cs:2.5, 0.1);

% Y-axis: tick + label C(0) at y=1
\draw[thick] (axis cs:0, 1) -- (axis cs:-0.06, 1);
\node[anchor=east, font=\normalsize] at (axis cs:-0.05, 1) {$\hat{y}_0$};

% Y-axis: tick + label C(1) at y=0.1
\draw[thick] (axis cs:0, 0.1) -- (axis cs:-0.06, 0.1);
\node[anchor=east, font=\normalsize] at (axis cs:-0.05, 0.1) {$\hat{y}_1$};

% X-axis: tick + label at x=2
\draw[thick] (axis cs:2, 0) -- (axis cs:2, -0.04);
\node[anchor=north, font=\normalsize] at (axis cs:2, -0.05) {$1$};

\node[anchor=north, font=\normalsize] at (axis cs:0, -0.05) {$0$};

\end{axis}
\end{tikzpicture}
    }
    \caption{
    Example perturbation curves in a \ac{MORF}, incremental deletion setting.
    Despite the dashed \textcolor{blue}{blue} curve representing an desirable case of faithfulness, \ac{AOPC} rewards the \textcolor{red}{red} curve with a perfect score (normalized AOPC of 1).
    However, the latter is a degenerate situation, whereby only the feature with the highest importance affects the model's prediction, while lower-importance features have no effect.
    On the other hand \ac{AOPC} should reward the blue case, abiding to the aforementioned assumption of proportionality.
    }
    \label{fig:auc}
\end{figure}

This point represents the fundamental criticism we raise to current \ac{FI} faithfulness metrics---the existing correlation-based and area-based metrics do not really measure faithfulness correctly.
Mir\'o-Nicolau \textit{et al.}~\shortcite{miro2025comprehensive} already showed, on simple experiments, that existing faithfulness metrics seem to be misaligned to the actual goal of measuring faithfulness.
In this section, we expand on their experiments by providing additional criticism on the current approaches.
For simplicity, we will work with an example of classification model, on which we evaluate explanations using guided perturbation fidelity obtained via \ac{MORF}, incremental deletion feature perturbation.
We additionally assume that the explanation only presents positive importance scores.
We expect the behavior of the curve to follow a trend similar to the one depicted in \Cref{fig:perturbation-curve}:
there should be a proportionality between the importance of features removed and the drop in confidence observed.
We can observe that, in this case, neither correlation-based, nor area-based metrics are aligned with measuring faithfulness since they break this assumption:
\begin{itemize}[label=$\blacktriangleright$]
    \item Pearson's correlation assumes a linear relationship between the two axes.
          Thus, a perfectly faithful explanation (see \Cref{fig:correlation}) would require a uniform drop in confidence for each iteration.
          This would mean that the contribution of each feature is the same, regardless of its importance.
    \item Nonlinear correlation metrics (e.g., Spearman's $\rho$) are just measuring the monotonicity of the trend, they do not consider its convexity.
          Thus, the proportionality assumption could be violated, since any monotonic trend would lead to a maximum faithfulness score, as depicted in \Cref{fig:correlation}.
          High nonlinear correlation is hence more of a \emph{necessary} property of a faithful explanation than a \emph{sufficient} one.
    \item \ac{AOPC} would require a perfectly faithful explanation to follow the trend highlighted in \Cref{fig:auc}.
          This conflicts with the expected trend from \Cref{fig:perturbation-curve} and violates the proportionality assumption:
          only the most important feature(s) would lead to a drop in confidence, and all the relevant features with lower importance would not have any effect on the model prediction.
          The only situation in which \ac{AOPC} would be a valid faithfulness metric is when the explanation is binarized, thus possibly leading to the behavior in \Cref{fig:auc}.
\end{itemize}

\begin{figure*}[t]
    \centering
    \begin{subfigure}[t]{0.33\textwidth}
        \centering
        \includegraphics[width=\textwidth]{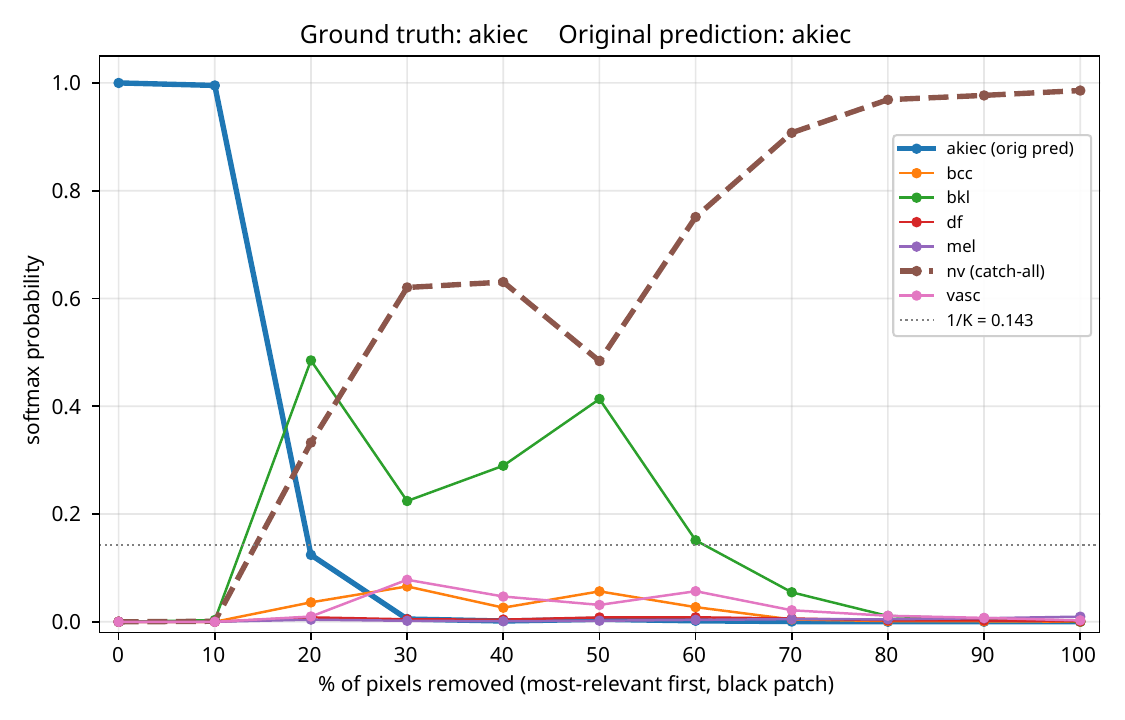}
        \caption{}
        \label{fig:skincancer_a}
    \end{subfigure}%
    \begin{subfigure}[t]{0.33\textwidth}
        \centering
        \includegraphics[width=\textwidth]{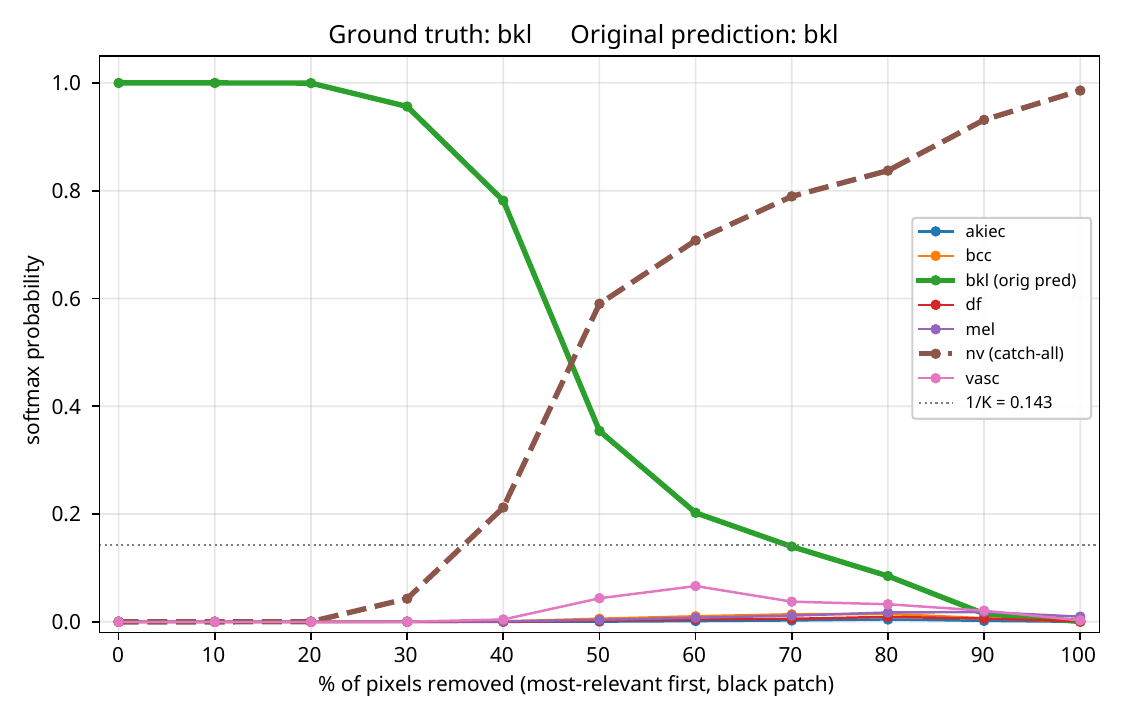}
        \caption{}
        \label{fig:skincancer_b}
    \end{subfigure}
    \begin{subfigure}[t]{0.33\textwidth}
        \centering
        \includegraphics[width=\textwidth]{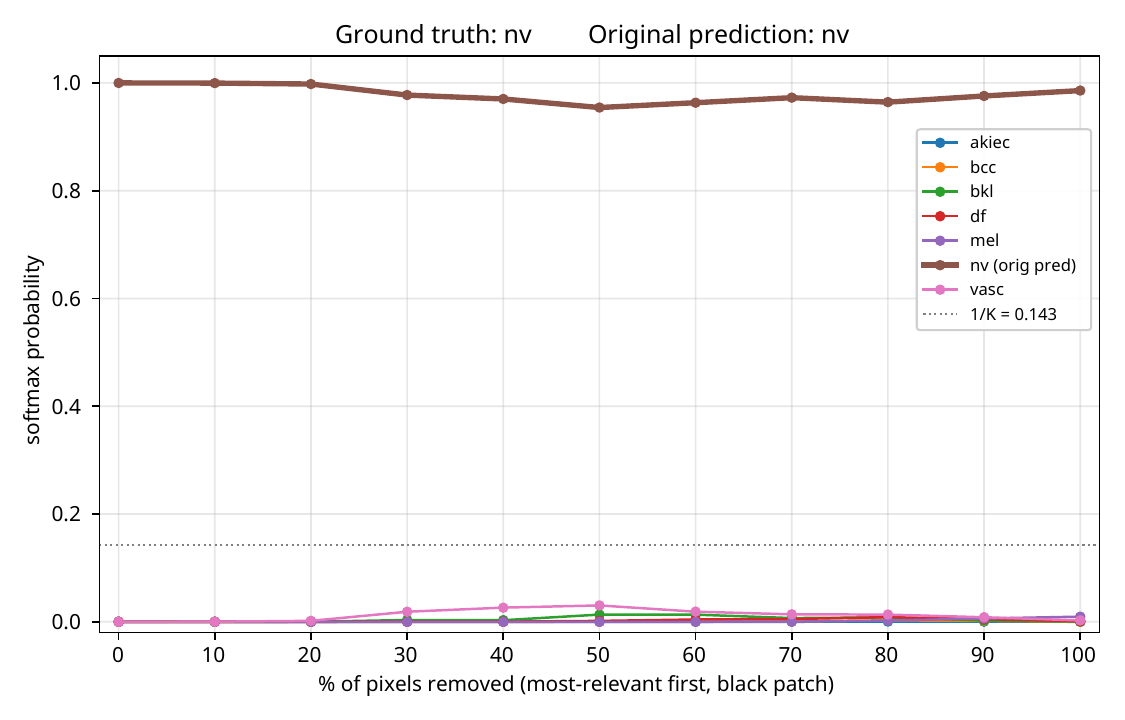}
        \caption{}
        \label{fig:skincancer_c}
    \end{subfigure}
    \caption{
    Example of perturbation curves obtained on a ResNet-18 \ac{ANN} \protect\cite{he2016deep} trained on the HAM10000 dataset \protect\cite{codella2019skin}.
    The dataset contains images from seven  classes of skin cancer (\texttt{akiec}=Actinic keratoses, \texttt{bcc}=basal cell carcinoma, \texttt{bkl}=benign keratosis-like lesions, \texttt{df}=dermatofibroma, \texttt{mel}=melanoma, \texttt{nv}=melanocytic nevi, \texttt{vasc}=vascular lesions).
    The plots are obtained by incrementally deleting features in \ac{MORF} order with black patches.
    In all cases, the model designs the \texttt{nv} class as a \emph{catch-all category}.
    (a) and (b) show images from ground truth classes different than \texttt{nv}, while (c) shows an example from the \texttt{nv} class, where the model confidence remains extremely high when the image is completely black.
    This demonstrates that faithfulness computation is flawed for models with poor \ac{OOD} capabilities.
    The $1/K$ line represents the optimal case for a well-calibrated model, which should output a flat distribution for completely black images. None of the three figures showcase anything close to this behavior.
    }
    \label{fig:catchall}
\end{figure*}

In addition, there is one more point to be made: the assumption of confidence drop in classification can only be reliably met when the faithfulness metrics are run on the predicted class.
In case the explanation is run on a non-predicted class, it is likely that the confidence would be very low, possibly close to 0.
In the ideal case of a perfectly calibrated model, removing important features would possibly cause an \emph{increase} in confidence prediction (towards $1/K$, with $K$ being the number of classes).
In the uncalibrated case, this confidence increase could be even higher, due to the existence of a so called \emph{catch-all category} ~\cite{miro2025comprehensive} that the model designates as a default class for \ac{OOD} examples.
This makes the current approaches infeasible for measuring the faithfulness of non-predicted classes.
\Cref{fig:catchall} shows a real-life example of perturbation curves in a model with poor \ac{OOD} capabilities, where one class is designated as a high-confident \emph{catch-all}.

\section{Our proposal for the direction of \ac{FI} faithfulness research}\label{sec:proposals}
Having outlined key issues with current faithfulness computation, we propose research directions towards more solid and trustworthy faithfulness evaluation. This aims to restore confidence in \ac{XAI} methods, which, per \Cref{sec:intro}, has declined due to unfulfilled promises regarding model interpretability.

\textbf{Proposal I -- faithfulness should be evaluated on well-calibrated models robust to \ac{OOD} data:}\quad
faithfulness relies on the assumption that models should reduce the confidence in their predictions as the data gets progressively perturbed, as explored in \Cref{sec:faithfulness}.
This collides with the fact that most \acp{ANN} are largely overconfident in their predictions, especially for what concerns \ac{OOD} data, as seen in \Cref{sec:criticism-ood}.
We hence posit that faithfulness should be computed on well-calibrated, \ac{OOD}-robust models.
In that regard, we believe the recent research on \ac{XAI}-aware model calibration \cite{sridhar2026mcal,cai2026rethinking} to be a valid direction, favoring a more reliable faithfulness computation.
Another underexplored line of research that needs to be expanded upon is \ac{XAI} for Bayesian neural networks \cite{valdenegro2024sanity}: these models tend to display a more natural robustness to \ac{OOD} data.
In addition to these benefits, calibration has other positive effects on the model trustworthiness:
a model whose uncertainty estimates are well calibrated enhances human oversight, allowing stakeholders to immediately get an indication on the possible (in)correctness of a prediction---outputs with low confidence are more likely to be inaccurate than highly-confident ones \cite{guo2017calibration}.

%\textbf{Proposal II: faithfulness for classification should be evaluated on the predicted classes}

\textbf{Proposal II -- faithfulness metrics for classification needs to consider the cases of non-predicted classes:}\quad
as outlined in \Cref{sec:criticism-formal-divergence}, faithfulness currently lacks an evaluation approach for explanations generated on non-predicted classes.
The generation of explanations on classes other than the predicted ones could be necessary in some cases, e.g., for contrastive explanations \cite{sarti-etal-2023-quantifying,zhang2025finer}.
Thus, we posit that approaches for faithfulness computation for classification tasks should generalize to all classes, not only the predicted one.
Until such approaches are formulated, we posit that practitioners should concentrate only on generating explanations on the predicted category. 

\textbf{Proposal III -- faithfulness for regression needs more research efforts:}\quad as discussed in~\Cref{sec:criticism-regression}, current faithfulness metrics are predominantly tailored to classification tasks and cannot be trivially adapted to regression settings without introducing conceptual problems. Unlike classification, regression requires metrics that assess whether \ac{FI} explanations correctly predict both the direction and magnitude of changes in predicted values. We argue that addressing this gap requires dedicated research efforts that explicitly account for the unique characteristics of regression tasks. Specifically, the community should develop formalizations of faithfulness metrics designed specifically for regression, rather than adapting classification-based approaches that rely on incompatible assumptions.
New metrics such as TIMING \cite{jang2025timing} represent a solid avenue, although the reliance on area-based metrics may still be problematic, due to the issues raised in the present paper.
We additionally highlight that this proposal may pose issues in conjunction with Proposal I, since calibration for regression models less well-established than for classification~\cite{levi2022evaluating}.

\begin{figure}[t]
    \centering
    \scalebox{0.7}{
        \begin{tikzpicture}

\begin{axis}[
  width  = 9cm,
  height = 9cm,
  xmin = 0, xmax = 1,
  ymin = 0, ymax = 1,
  xtick = {0, 0.2, 0.4, 0.6, 0.8, 1.0},
  ytick = {0, 0.2, 0.4, 0.6, 0.8, 1.0},
  xlabel = {Predicted confidence},
  ylabel = {Observed accuracy},
  label style   = {font=\small},
  tick label style = {font=\small},
  grid = major,
  grid style = {gray!20, very thin},
  axis line style = {thick},
  clip = false,
  legend style={at={(0.03,0.9)},anchor=west}
]

%% -------------------------------------------------------

%% -------------------------------------------------------
%% 3.  Perfect-calibration diagonal
%% -------------------------------------------------------
\addplot [
  color     = gray,
  thick,
  dashed,
  line width = 1.5pt,
] coordinates {(0,0) (1,1)};
\addlegendentry{Perfect calibration};

%% -------------------------------------------------------
%% 4.  Overconfident model curve
%%     accuracy < confidence  =>  points lie BELOW diagonal
%%     Bin centres: 0.1, 0.3, 0.5, 0.7, 0.9
%%     Model accuracy:  0.05, 0.15, 0.30, 0.45, 0.65
%% -------------------------------------------------------
\addplot [
  color      = black,
  thick,
  line width = 2pt,
  mark       = *,
  mark size  = 3pt,
] coordinates {
  (0.1, 0.05)
  (0.3, 0.15)
  (0.5, 0.30)
  (0.7, 0.45)
  (0.9, 0.65)
};
\addlegendentry{Model (overconfident)};

%% -------------------------------------------------------
%% 5.  Vertical dashed lines: model point  ->  diagonal
%%     On the diagonal accuracy = confidence, so the top
%%     of each line is at (conf, conf).
%% -------------------------------------------------------
\foreach \conf/\acc in {0.1/0.05, 0.3/0.15, 0.5/0.30, 0.7/0.45, 0.9/0.65}{
  \addplot [
    color      = {rgb,255: red,0; green,0; blue,255},
    %thin,
    dotted,
    line width = 0.8pt,
  ] coordinates {(\conf, \acc) (\conf, \conf)};
}

%% -------------------------------------------------------
%% 6.  Region labels
%% -------------------------------------------------------
\node [
  font  = \small\bfseries,
  color = {rgb,255: red,160; green,80; blue,0},
] at (axis cs: 0.75, 0.25) {Overconfident};

\node [
  font  = \small\bfseries,
  color = {rgb,255: red,0; green,100; blue,180},
] at (axis cs: 0.28, 0.74) {Underconfident};

\end{axis}

%% -------------------------------------------------------
%% 7.  legend

legend style={at={(0.03,0.5)},anchor=west}

\end{tikzpicture}
    }
    \caption{Example of calibration plot for a toy classification model.
    The model is evaluated on a validation dataset.
    The data are then split into bins depending on the confidence and the average accuracy per bin is computed and plotted (solid black line).
    The \ac{ECE} is calculated as the magnitude of differences (dotted blue lines) between these points and the diagonal (gray dashed line), which represents perfect calibration where confidence equals accuracy.}
    \label{fig:ece}
\end{figure}
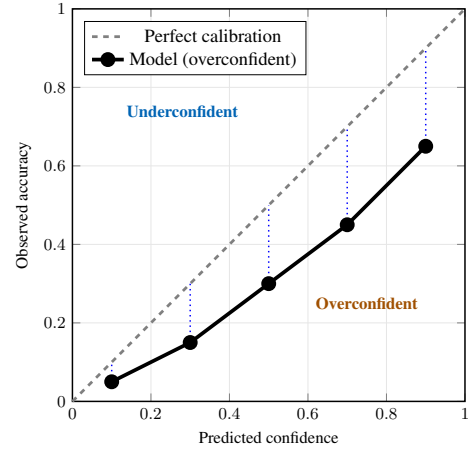

\textbf{Proposal IV -- faithfulness metrics should be more aligned to their goal:}\quad
with regards to what has already been outlined in \Cref{sec:criticism-formal-divergence}, we believe the current approaches based on correlation and area-based metrics (and related work-arounds) should be scraped, favoring instead approaches that consider what are the \emph{expectations} from a faithful explanation, and that reliable metrics be developed abiding to these expectations.
We consider the following points:
\begin{enumerate}[label=\roman*.]
    \item The x-axis of the perturbation curve should represent the importance of the features removed, while many approaches use the iteration number instead.
    The latter does not connect to a direct value of \ac{FI}, which does not allow to get a sense of the prediction change proportional to said importance.
    \item If an optimal shape of the prediction curve is established (e.g., similar to the one in \Cref{fig:perturbation-curve}), then an inspiration for the faithfulness computation could be coming from the field of uncertainty estimation.
    Specifically, the \ac{ECE} already offers an example of a metric considering a difference with an expected curve shape---see \Cref{fig:ece}.
    We stress out that currently there does not exist an agreed-upon definition of \emph{optimality} for a perturbation curve, and we posit more research should be devoted to eliciting this.
    The choice of perturbation and phenomena such as gradient starvation (see \Cref{sec:criticism-formal-divergence}) may pose a big hurdle in determining an optimal shape of the curve.
    %\item Exp regression? 
    \item Since faithfulness metrics should foster model trustworthiness, these should necessarily be relative in order to allow for direct interpretation of the faithfulness value.
    Absolute metrics calculated, e.g., on the output logits, such as in the case of IROF \cite{rieger2020irof} 
    do not offer a relative value (e.g., between 0 and 1) and logits can easily differ between models, making comparisons impossible and faithfulness scores uninterpretable.
    %If a relative value is not possible, at least the faithfulness value should be comparable, e.g., to the unit of measurement of the data or to the loss/error, in order to give the model stakeholders a mean to  \emph{understand} what this value means.
\end{enumerate}

%Talk also about replacing x-axis in perturbation curve with importance of removed features, not iteration number (does not align with the proportionality assumption).
%Talk also about the fact that metrics should necessarily be relative, since absolute metrics do not allow for comparisons.
%Talk about insights from uncertainty quantification, where the ECE is calculated as a difference w.r.t. the optimal behavior

\section{Contextualizing Faithfulness Assessment}\label{sec:related}

Issues with faithfulness metrics are well-known in the \ac{XAI} literature.
Ever since the still-used pixel/feature flipping experiment \cite{bach2015pixel}, subsequent methods have tried to address its shortcomings, mainly on \ac{OOD} data created in the perturbation process:
we have cited the examples of ROAR \cite{hooker2019benchmark}---which proposes a retraining process---and the derivative ROAD \cite{rong2022consistent}---which instead proposes blurring to avoid the expensive retraining.
More recently, works like F-fidelity \cite{zheng2024ffidelity} have proposed the usage of augmentative finetuning to curb the \ac{OOD} issues.
However, there are papers which have targeted a more comprehensive critical approach to faithfulness evaluation:
the previously cited work by Mir\'o-Nicolau \textit{et al.}~\shortcite{miro2025comprehensive} indicates how the main approaches seem to be misaligned with the goal of measuring faithfulness in models trained on simple tasks, where the important features are known in advance.
The authors frame their findings on a critical level, not listing concrete proposals for future directions.
Additionally, Chan \textit{et al.}~\shortcite{chan2022comparative}, in their comparative study, notice inconsistencies between different faithfulness metrics on natural language processing tasks.
They argue in favor of correlation-based metrics; however, they miss the issues behind these approaches which we instead outline in \Cref{sec:criticism-formal-divergence}.
Seth and Sankarapu \shortcite{seth2025bridging} highlight the importance of a sound evaluation of \ac{XAI} and criticize the current field because of 
\begin{enumerate*}[label=(\roman*)]
    \item fragmentation---a lack of standard in metrics---and
    \item manipulability---metrics are not robust and can be manipulated easily.
\end{enumerate*}
However, their work is designed at a governance level and does not address the technical details behind these shortcomings.
Finally, the previously mentioned Haufe \textit{et al.}~\shortcite{haufe2026explainable} also address \ac{XAI} shortcomings.
Their perspective encompasses several points, including the lack of formalization in \ac{XAI}, the fragility of the current metrics for \ac{XAI} evaluation, the lack of verification that explanations abide by the formal requirements, and the scarcity of real-world data for performing empirical evaluations.
The authors also mention that current \ac{XAI} tools and faithfulness metrics do not take into consideration the data generating distribution; however, they do not propose how to integrate this into existing methods.
Our proposals go in the direction highlighted by the authors;
specifically, we also call for higher formality in \ac{XAI} evaluation, e.g., by soliciting research to define a desired form of perturbation curve (Proposal IV) and by posing more stringent requirements on the model calibration (Proposal I).

\section{Discussion and conclusion}\label{sec:discussion}

In the present paper, we have summarized several issues connected with the current metrics for the computation of faithfulness of \ac{FI}.
Faithfulness is defined as the level of alignment between the \ac{FI} explanation and the internals of the model on which said explanation is generated.
We highlighted how the main metrics approach the computation from the process of \emph{data intervention} \cite{dembinsky2026unifying}: perturbing specific features of a given input should yield variations in the model output which are proportional to the importance of the features.
This is often done in an iterative fashion, by incrementally \emph{removing} (or \emph{revealing}) features by means of perturbations operated in a \ac{MORF} or \ac{LERF} fashion.
The output of the model is tracked for each perturbed input and then plotted against the fraction of features removed, as shown in \Cref{fig:perturbation-curve}. 
Faithfulness is then summarized with a scalar by either computing the correlation of the perturbation curve or its area under (or over) the curve.

We posit that these approaches present several flaws. Some of these criticisms are novel, while some others are already well known in the literature.
We present five main shortcomings:
\begin{enumerate*}[label=(\arabic*)]
    \item disproportionate focus on classification for faithfulness research,
    \item conflation of faithfulness with other \ac{XAI} evaluation facets,
    \item necessity of an arbitrary baseline for simulating feature removal,
    \item emergence of \ac{OOD} data (which most \acp{ANN} handle poorly) as effect of the perturbation, and
    \item misalignment between the main metrics and the proper computation of faithfulness.
\end{enumerate*}
These drawbacks severely limit the assessment of faithfulness---a metric whose importance is vital in \ac{XAI} evaluation, since it essentially quantifies the \emph{level of approximation} of an explanation with regard to the model, and is hence fundamental for fostering reliability of \ac{FI} as the main \ac{XAI} task.
Increasing the trustworthiness of \ac{XAI} is hence necessary, in a time in which, after roughly a decade of research, the field is being severely criticized for its shortcomings in fulfilling its initial commitments in improving model transparency for its users \cite{afroogh2026beyond}.
We conclude our paper by proposing four avenues for faithfulness research, namely:
\begin{enumerate*}[label=(\Roman*)]
    \item contemplate model calibration when \ac{FI} needs formal evaluation,
    \item for classification, consider that explanations can be generated either on predicted or non-predicted classes (while current methods concentrate solely on predicted classes),
    \item direct more effort to faithfulness for regression, and
    \item align metrics to the actual goal of computing faithfulness.
\end{enumerate*}
We hope our manuscript can provide a clear and comprehensive summary of the current problems with the computation of faithfulness, at the same time giving directions to practitioners and researchers on how the research in the field can be improved.
We close our paper by shifting back to \Cref{sec:intro}---we posit Afroogh \textit{et al.}'s~\shortcite{afroogh2026beyond} point on infidelity to be inaccurate: it is not that \ac{XAI} is unfaithful, but we are currently lacking a principled way for assessing faithfulness.
Even though a more formal approach to faithfulness will not solve all of the shortcomings of \ac{XAI}, as outlined by Haufe \textit{et al.}~\shortcite{haufe2026explainable}, this will still help with bolstering its credibility.

\textbf{Limitations}:
despite providing a comprehensive critical outline of the current metrics for \ac{FI} faithfulness, our paper does not consider other popular \ac{XAI} tasks, such as counterfactual examples or concepts, that all require faithfulness computation \cite{dembinsky2026unifying}. 
Concepts, for instance, present similar challenges to \ac{FI} in this regard.
Additionally, if we exclude the example from \Cref{fig:catchall}, the present paper lacks experiments that summarize the issues we highlight in \Cref{sec:faith-issues}, although many of these problems are already well-known in the literature.
Finally, some of our directions from \Cref{sec:proposals} (e.g., Proposals II and III) act more as recommendations for research and do not contain any concrete idea on how the limitations should be tackled.
Similarly, the limitation on the arbitrary nature of perturbation baselines from \Cref{sec:criticism-baseline} is not tackled by our proposals and remains currently unresolved. 
Notwithstanding these limitations, we believe our position paper still effectively summarizes shortcomings of current approaches and provides valuable insights for directing future research on faithfulness. 

\subsection*{Acknowledgment}\label{ack}
This work is partly supported by BMFTR (Federal Ministry of Research, Technology and Space) in DAAD project 57616814 (SECAI, School of Embedded Composite AI, https://secai.org/).
\bibliographystyle{named}
\bibliography{ijcai26}

\end{document}